# Applying unsupervised keyphrase methods on concepts extracted from discharge sheets

**Hoda Memarzadeh**

Department of Electrical and Computer Engineering, Isfahan University of Technology, Isfahan 84156-83111, Iran. Electronic address: h.memarzadeh@ec.iut.ac.ir

**Nasser Ghadiri**[1]

Department of Electrical and Computer Engineering, Isfahan University of Technology, Isfahan 84156-83111, Iran. Electronic address: nghadiri@iut.ac.ir

**Matthias Samwald**

Institute for Artificial Intelligence, Center for Medical Statistics, Informatics, and Intelligent Systems, Medical University of Vienna, Vienna, Austria. Electronic address: matthias.samwald@meduniwien.ac.at

**Maryam Lotfi Shahreza**

Department of Computer Engineering, Shahreza Campus, University of Isfahan, Iran. Electronic address: m.lotfi@shr.ui.ac.ir

**Abstract**

Clinical notes containing valuable patient information are written by different health care providers with various scientific levels and writing styles. It might be helpful for clinicians and researchers to understand what information is essential when dealing with extensive electronic medical records. Entities recognizing and mapping them to standard terminologies is crucial in reducing ambiguity in processing clinical notes. Although named entity recognition and entity linking are critical steps in clinical natural language processing, they can also result in the production of repetitive and low-value concepts. In other hand, all parts of a clinical text do not share the same importance or content in predicting the patient's condition. As a result, it is necessary to identify the section in which each content is recorded and also to identify key concepts to extract meaning from clinical texts. In this study, these challenges have been addressed by using clinical natural language processing techniques. In addition, in order to identify key concepts, a set of popular unsupervised key phrase extraction methods has been verified and evaluated. Considering that most of the clinical concepts are in the form of multi-word expressions and their accurate identification requires the user to specify n-gram range, we have proposed a shortcut method to preserve the structure of the expression based on TF-IDF. In order to evaluate the pre-processing method and select the concepts, we have designed two types of downstream tasks (multiple and binary classification) using the capabilities of transformer-based models. The obtained results show the superiority of proposed method in combination with SciBERT model, also offer an insight into the efficacy of general extracting essential phrase methods for clinical notes.

# 1 Introduction

An electronic medical record (EMR) contains information about a patient's health history, such as diagnoses, medicines, tests, allergies, immunizations, and treatments. This information includes structured and unstructured data. The structured data, including laboratory tests, medications, diagnostic codes, and prescriptions, are recorded in patient files using standard coding systems, such as ICD10[2], RxNorm[3], and LOINC[4]. Unlike structured data, unstructured data are free-form texts written by service providers. An EMR stores unstructured data, such as patient history and explanations of their illness, textually. Processing these texts, however, could be challenging due to syntactic differences, word choice, abbreviations, acronyms, compounds, negation expressions, speculative cues, factuality expressions, and spelling errors [1]. Nevertheless, it is crucial to integrate clinical, imaging, and molecular profiling data to detect complicated diseases and provide accurate diagnoses [2].

Accordingly, unstructured data processing has become one of the most researched topics, given its complexity. Overcoming the challenges of processing clinical notes requires named entity recognition (NER) and named entity disambiguation or entity linking. The NER identifies and aligns biomedical entities to knowledge bases, enhancing decision-making efficiency [3] and reducing the diversity of clinical terms. A key goal of entity linking is to give each entity in a text a unique identifier. Through entity linking, entities extracted from the text are mapped to unique identifiers from a knowledge base, for example, the Unified Medical Language

---

[1] Corresponding author. Tel.: +98-313-391-9058; fax: +98-313-391-2451; e-mail: nghadiri@iut.ac.ir
[2] International statistical classification of diseases and related health problems
[3] The normalized naming system for generic and branded drugs
[4] Logical Observation Identifiers Names and Codes

System (UMLS). The problem that arises when employing NER and entity linking methods is that the extracted concepts can be repetitive and may not have a differential value.

Keyphrase (or keyword) extraction is fundamental in information management systems. It is the process of extracting keyphrases from a document, i.e., a set of phrases consisting of one or more words that are considered meaningful and representative [4]. Although various supervised and unsupervised keyphrase extraction methods have been proposed in the previous researches [5], in practice there still occur special issue regarding multi-word phrases and specifying proper n-gram range. A word n-gram range lets users decide the length of the sequence of consecutive words that should be extracted from a given text. However, users usually do not know the optimal n-gram range and therefore have to spend some time experimenting until they find a suitable n-gram range. Furthermore, this means that grammatical sentence structures are not considered at all. This leads to the effect that even after finding a good n-gram range, the returned key phrases are sometimes still grammatically not quite correct or are slightly off-key. To address the issues mentioned above we proposed simple method that use unique code of concepts in TF-IDF algorithm.

To implement the algorithm, we use discharge sheet as the most important clinical note in EMR and consists of several components or sections [6]. In the first step of this study, the discharge sheet of each patient was split into inner sections.

By selecting a subset of sections with the phicysion's opinion, in the next step, the content of the selected sctions will be converted into a set of clinical concepts during the NER and entity linking process. The concept set which will be used to find the key concepts by the proposed method, which is a kind of TF-IDf implementation.

We compare our result with unsupervised keyphrase extraction techniques (statistical, graph-based, and deep learning techniques). In order to evaluate the proposed method, the output of all algorithms, which includes the subset of selected concepts by each algorithm, is converted into a string and using transformer-based representation models apply in two binary and multiple downstream tasks, for mortality and diagnoses prediction, respectively.

The contribution of this paper lies in (i) Applying unsupervised keyphrase extraction methods based on extracted concepts from discharge sheets and comparing the result through classification downstream tasks, (ii) Examining the accuracy of general and specific language transformer models based on different inputs provided by keyphrase extraction methods (iii) Analyzing the applicability of general keyphrase extraction methods to clinical concepts.

The remainder of the paper is organized as follows. Section 2 describes the clinical NLP, clinical transformer-based representation model, and unsupervised keyphrase extraction methods assessed. The proposed framework is presented in Section 3. Section 4 presents the results of the comparative assessment of the selected methods. The discussion concludes in Section 5, followed by concluding remarks and future work directions.

## 2  Related Work

Entity linking and NER reduce ambiguity, but they also produce repetitive and insignificant concepts, this problem is evident in the processing of clinical texts due to the variety of specialized expressions. Therefore, in this research, we have tried to evaluates keyphrase extraction methods based on clinical concepts using representation vectors. Considering that in this study several fields of study such as clinical natural language processing (cNLP), Clinical transformer-based representation models, and keyphrase extraction methods have been used to present the proposed method and evaluations, in this part we will review some of the developments and studies related to these fields:

### 2.1  Clinical natural language processing (cNLP)

A critical step in natural language processing (NLP) is entity recognition. This process can be rule-based or a combination of methods in which machine learning models are used to reinforce rules [7]. In recent years, the machine learning method has witnessed significant growth [8], [9]. The main advantage of machine learning approaches over rule-based ones is their higher flexibility. Meanwhile, they require large quantities of word-level annotations, which are costly and impossible to obtain within biomedical settings [10]. The ScispaCy is an instance of the rule-based technique [11]. A custom tokenizer adds tokenization rules onto SpaCy's rule-based tokenizer [12]. After NER and to reduce ambiguity, extracted entities from the text were mapped to corresponding unique ids from a target knowledge base [13]. In ScispaCy, the target knowledge base could be selected from UMLS, MeSH, RxNorm, GO, or HPO [11].

### 2.2  Unsupervised keyphrases extraction methods

Despite the added value achieved through NER and entity linking, there is always the issue that the set of concepts extracted includes concepts that do not have differentiating characteristics. In other words, more valuable concepts need to be separated from other extracted concepts [14]. Unsupervised keyphrases extraction methods follow a standard three-step methodology according to [3], [15]. The first step is to filter unnecessary lexical units from the input text using heuristics. In order to rank the units, some syntactic/semantic relationships with other candidate units are considered. Based on the ranked list of candidate words, key phrases are extracted. This section presents the most prominent statistical, graph-based, and deep learning keyphrases extraction methods.

Statistical Methods: Based on the statistical characteristics of the words in the text, such as their repetitions in the text or their position in the text, statistical methods calculate a score for each phrase. A statistical method uses a different approach for calculating scores. In terms of statistical methods, we can consider TF-IDF [16], YAKE [17], and KP-miner[18] the most important.

Graph-Based Methods: By using a graph-based algorithm based on the assumption that more connections equal more important candidate words, the keyphrase extraction task is transformed into a graph sorting problem. Several graph-based models, such as PositionRank [19], TextRank [20], SingleRank [21], and MultipartiteRank [22], have been proposed to identify meaningful co-occurrence information between words.

Deep Learning Methods: The methods referred to as Deep Learning Methods are based on the use of Language Model. This category of methods involves learning rich representations of keyphrases in a language model to improve keyphrase extraction and generation performance. In this category of methods, language models are used to learn rich representations of keyphrases for improved keyphrase generation and extraction [15]. The methods like KeyBERT [23], Key2vec [24] are in this category.

The extraction of keywords directly from clinical texts has been investigated in previous studies[25], among which we can refer to - The difference of the proposed method is that in this study first the primary texts have been converted into a set of cited concepts and then key concepts have been extracted.

## 2.3 Clinical transformer-based representation models

While cNLP focuses on concepts, representation learning can help build complex semantic relationships between extracted concepts by representing their semantics in one unified space [26]. Over the years, researchers have developed numerous text representation techniques, which started with one-hot and bag-of-words models and expanded to RNN-based and transformer-based models. Given the general success of transformer-based pre-trained language models, many such models have been developed by pre-training on biomedical corpora. In recent years, we have seen the development of textual biomedical pre-trained models. These models can be classified from different perspectives such as data resource and pre-train strategy, base model and etc. [27]. The methods like BioBERT[28], SciBERT[29], Clinical BERT[30], UmlsBERT[31], PubMedBERT [32] are examples of biomedical pre-trained models.

## 3 Methodology

Our goal in this paper is to present an optimal framework for getting from raw clinical text to a machine-processable representation. In this framework, we try to provide the best sequence of different approaches in each step. A workflow in Figure 1, uses the embedding vector to predict patient diagnoses based on the representation of key concepts extracted from clinical texts. Because clinical texts contain specialized terms that are recorded in different EMR forms, we did not perform keyphrase extraction methods directly on the raw clinical texts, but first, we mapped the raw clinical notes to clinical concepts by preprocessing them through several steps. Using the concepts obtained from the first stage, keyphrase extraction methods are applied. The extracted keyphrases are represented by pre-trained biomedical language models. After that, each keyphrase extraction method is evaluated using multi-label classification. In the following, the details of each step will be explained.

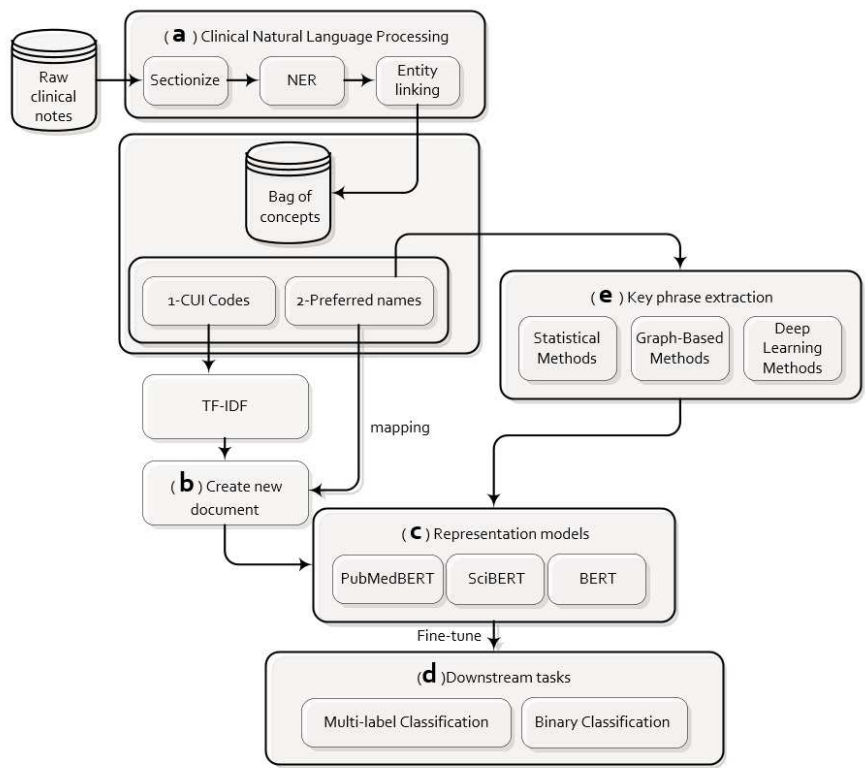

**Fig. 1.** Overview of the proposed method. First of all, in step (a), discharge sheet for each patient is taken from the dataset and trough clinical natural language processing pipeline convert to bag of concepts. In this step, two properties of concepts are saved. 1- CUI concepts codes, 2- Standard expressions corresponding to concepts (preferred names). Afterward in section (b), dataset of CUI concepts codes processed by the TF-IDF algorithm to detect ones with scores above threshold. Then the selected concepts mapped to the preferred names to creating a dataset. In step (c), three clinical transformer-based representation models fine tune on generated dataset for use in downstream tasks (step (d)). In step (e) we compare our proposed method with different categories of keyphrase extraction methods that directly run on second dataset (Dataset generated from expressions corresponding to concepts).

## 3.1 Clinical natural language processing (cNLP)

First step of the proposed workflow is clinical natural language processing. The input of this step is a dataset of discharge notes. Discharge sheet is an essential document in an electronic medical record, Each of the discharges notes, consists of several components or sections [6]. However, there is significant variation in sections and the descriptive phrases in section headers. This challenge in the NLP of clinical notes is named flexible formatting [33]. Selected sections are: past medical history, history of present illness, chief complaint, family history, physical exam. These sections were selected with the doctor's opinion.

After dividing each discharge note to constituent parts sections by medspaCy [34], we follow the next step, which is named-entity recognition (NER). In NER step, biomedical entities were identified. For NER, we use the en_ner_bc5cdr_md model, a spaCy NER model trained on the BC5CDR corpus. The BC5CDR corpus consists of 1500 PubMed articles with 4409 annotated chemicals, 5818 diseases, and 3116 chemical-disease interactions. The en_ner_bc5cdr_md model covers two types of entities: DISEASE and CHEMICAL. In the next step, all extracted entity types are used. Also, the phrases that appeared negatively in the notes and were identified by Negspacy [35] were added by a prefix "NOT".

The next step after NER is entity linking. A key goal of entity linking is disambiguation and reducing clinical terminology diversity. By using entity linkage, entities extracted from the text can be mapped to corresponding unique identifiers from a knowledge base, such as the Unified Medical Language System (UMLS). The Scispacy python library [11] was used to identify entities in selected sections, and the detected entities were mapped with the UMLS Metathesaurus.

The output of these steps is a repository of Concepts including CUI codes and related preferred named to each concept.

## 3.2 Creating new document by TF-IDF

As shown in section (b) in Figure 1, we use the CUI codes for extracting key concepts by TF-IDF algorithm. Based on the frequency of a CUI code and how many other documents include it, TF-IDF[16] calculates a score for each CUI concept by equation (1).

$$TF-IDF_t = TF \times \log(\frac{|D|}{|d \in D : t \in d|}) \quad (1)$$

Where $TF-IDF_t$ is the homonym score for CUI code $t$, $TF_t$ is its term frequency, $|D|$ the number of documents, and $|d \in D: t \in d|$ the number of documents where $t$ is included. The selected CUIs were converted to their preferred names in the knowledge base. This was done using a dictionary that had been produced from all positive and negative concepts in the corpus. This was to preserve the phrase attributed to the concepts whose standard names consisted of several words, some of which were common between different concepts. In addition, negative expressions would appear in the final output with a "NOT" in front of their name. By concatenating the preferred names of selected concepts, a new document creates for each patient.

## 3.3 Clinical transformer-based representation models

To check the performance of keyphrase extraction based on clinical concepts codes, the previous stage's output was used in binary and multi-label classification downstream. Fine-tuned document representations were used to design the classification model. The clinical transformer-based representation models used in this step describe bellow:

1. The BERT-BASE[36] model that has been trained on the Wikipedia and book corpus databases but has not been fine-tuned on clinical data and can therefore be considered a general model.
2. The SciBERT model[29]. Among the first BERT-based models, SciBERT built its vocabulary, SciVocab. In this corpus, 1.14 M papers were randomly selected from Semantic Scholar with the same number of subwords as BERT_BASE's vocabulary (30K). 82% of the papers are in the biomedical domain, and 18% are in computer science. Only 42% of the text used in these scientific domains overlaps with BERT's vocabulary, indicating how different it is from public domain texts. Following the creation of SciVocab in both case and uncased versions, the model is pre-trained using scientific datasets using the same architecture as BERT_BASE. Because SciBERT uses a different vocabulary than BERT, it does not use BERT's weights as initialization. In addition to text classification, sequence labeling (e.g., NER), and dependency parsing, SciBERT uses the same architecture, hyperparameters, and optimization as BERT_BASE.
3. The PubMedBERT model that has its own vocabulary, which contains more words specific to the biomedical field than SciBERT, has been compiled from 30 million PubMed abstracts [32].

In the initial dataset, an array of diagnoses was created for each patient. A patient's first three final diagnoses were used after mapping to HCUP Clinical Classifications Software (CCS) for ICD-9-CM [37].

## 3.4 Downstream task

## 3.5 Unsupervised keyphrases extraction

Although named entity recognition and entity linking are critical steps in clinical natural language processing, they can also result in the production of repetitive and low-value concepts. In the second stage we extract the essential concepts by applying three categories of keyphrase extraction methods. These selected methods are as below:

### 3.5.1 Statistical Methods

Based on the statistical characteristics of the words in the text, such as their repetitions in the text or their position in the text, statistical methods calculate a score for each phrase. We compared the proposed method with one of prominent algorithms of this group, which is YAKE [17]. In addition to term frequency, this algorithm considers context and terms scattered throughout the document. As soon as the text has been split into individual terms, YAKE calculates a score for each term t. There are five metrics used to calculate this score include casing aspect of a term, the positional of a term, term frequency normalization, term relatedness to context and the number of appearing a term in different sentences. YAKE allows only to set a max n-gram size and not a min size. The first 20 phrase extracted by the YAKE algorithm are considered as output.

### 3.5.2 Graph-Based Methods

By using a graph-based algorithm based on the assumption that more connections equal more important candidate words, the key phrase extraction task is transformed into a graph sorting problem. We compared the proposed method with two examples of prominent algorithms of this group, which are PositionRank [19] and MultipartiteRank [22].

A PositionRank algorithm graph is formed by representing nodes as words and edges as the co-occurrence counts of the words in the document. Furthermore, the PositionRank algorithm weights each word based on its position in the document. Hence, the words appearing early in the document are given a higher weight than those appearing later[19].

A directed graph is formed using candidate keyphrases instead of words in the MultipartiteRank algorithm. Additionally, a multipartite graph is constructed where two candidate keyphrases (nodes) are connected only if they are related. Topic identification can be performed using any existing method, such as Latent Dirichlet Allocation (LDA). The first candidate keyphrases in each topic are adjusted in weight by multiplying the inverse of their position number in the document. In both works (PositionRank and MultiPartiteRank), the authors observe that the quality of the extracted keyphrases is better than other approaches that do not leverage positional information [22]. In the implementation of PositionRank and MultipartiteRank, it is possible for the user to choose n-gram.

### 3.5.3 Deep Learning Methods

The methods referred to as Deep Learning Methods are based on the use of Language Model. This category of methods involves learning rich representations of keyphrases in a language model to improve keyphrase extraction and generation performance.

We compared the proposed method with two examples of prominent algorithms of this group, which are KeyBERT [23] and Key2vec [24].

The KeyBERT [23] algorithm, proposed by Grootendorst, utilizes pre-trained BERT-based word embedding models to enhance the quality of keyphrase extraction. A document-level representation is obtained using BERT first by extracting embeddings from documents. In the next step, N-gram word embeddings are extracted. Cosine similarity determines which words/phrases are most similar to the document. By identifying the most similar words, it would be possible to determine which words describe the entire document accurately.

Key2vec [24] combines the advantages of distributed phrase representations and PageRank-based ranking methods. Representation methods such as Fast text [38] represent the candidate phrases. Afterward, it forms a graph of candidate phrases whose edges are weighted according to semantic similarity. In the personalized PageRank algorithm, the nodes are ranked as usual. Compared to the existing state-of-the-art methods, the authors observed increased precision and recall when they applied the algorithm for keyphrase extraction from scientific documents.

At the end of this stage, the initial dataset which included the bag of concepts is converted into separate datasets. A sequence of concepts is created for each method and for each patient using expressions corresponding to key concepts in each file.

# 4 Evaluation

We used the Python programming language to implement and evaluate the selected keyphrase extraction methods. The complete codes, datasets, and evaluation results of our experiments are freely available on GitHub[5].

## 4.1 Datasets

MIMIC-III is a freely accessible database that includes de-identified health-related information on more than 46,520 patients admitted to a medical center between 2001 and 2012 [39]. The number of registered admissions for all patients is 58,976. We extracted data on recorded admissions for patients aged 18 to 99, leading to 48,058 admissions. Among the clinical forms registered for the patient, we used the discharge summary form. The number of 26,341 discharge summary forms was extracted after removing duplicates. These forms were first processed to identify the specific sections by the python tool MedspaCy [34]. The number of unique concepts extracted at the end of this stage, including those used negatively, includes more than 17 million concepts. We use 80% data for training and the remaining 10% data for validation and 10% for test.

---

[5] https://github.com/HodaMemar/A3.

## 4.2 Experimental Setup

This paper uses the Google Colab GPU runtime, which offers 12GB of memory and a Tesla K80 GPU processor with 2,496 CUDA cores. For each BERT-based model, we set the batch size to 32, the maximum sequence length to 128, and the learning rate to 2e-5. In order to avoid overriding the pre-trained weights too much, we use a small learning rate. We set the number of epochs to 5. Using BERT-based model as a baseline representation model and the SciBERT and the PubMedBERT as clinical teransoformer-based models, we will predict one or more labels for a given piece of generated text. For Clinical transformer-based representation models we use Huggingface [40].

As a linear layer is added on top of the base model, it produces a tensor of shapes (batch_size, num_labels), which indicates the unnormalized scores for a number of labels for each example. The loss function in "multi_label_classification" problem is BCEWithLogitsLoss.

For evaluation of multi-label classification we follow Longformer-multilabel-classification[6]. In this process the Longformer [39] architecture adopt to a multi-label setting for evaluation. Sigmoid function applied on predictions that are of shape (batch_size, num_labels); next, threshold used to turn them into integer predictions, and finally, metrics compute. In order to check the quality of the model, F1 and AUC (Area under the Curve) ROC (Receiver Operating Characteristics) curves have been used in the evaluations.

All the tools and python library are listed in Table 1:

**Table 1. The libraries used in the implementation of different stages of the process**

| Category | Methods | Tools/Python libraries/Setting | Ref |
|---|---|---|---|
| Clinical natural language processing (cNLP) | Section detection | MedspaCy/ NLP.pipe_name : 'medspaCy_sectionizer' | [34] |
| | Negation detection | Negspacy | [35] |
| | NER/Entity linking | ScispaCy/ en_ner_bc5cdr_md model | [11] |
| Statistical Methods | TFIDF | Scikit-Learn's TfidfVectorizer, norm='l2' , threshold score : >0.2, >0.4, >0.6 | [41] |
| | YAKE | pke - python keyphrase extraction | [42] |
| Graph-Based Methods | MultipartiteRank | | |
| | PositionRank | | |
| Deep Learning Methods | Key2vec | https://github.com/MarkSecada/key2vec/ | [24] |
| | KeyBERT | https://pypi.org/project/keybert/ | [23] |
| Clinical transformer-based representation models | BERT | "bert-base-uncased" | [43] |
| | PubmMedBert | 'microsoft/BiomedNLP-PubMedBERT-base-uncased-abstract-fulltext' | [32] |
| | SciBert | 'allenai/scibert_scivocab_uncased' | [29] |

## 4.3 Evaluation result

The evaluations are designed to answer the following research questions:
1. Are the general keyword extraction methods effective for notes included standard clinical concepts?
2. How did the representation model affect the quality of the multi-label classification model?
3. Can the direct use of concept codes instead of their corresponding expressions in the phase of extracting key concepts produce better results?

For answering above questions, we present the results of implementing binary classification model for mortality prediction and also multi-label classification model to predict patients' diagnoses. In both scenarios by fine-tuning representation models to the key concept extraction method outputs, embedding vector datasets are generated and apply for classification.

Table 2 includes the results of accuracy metrics in mortality prediction (binary classification).

---
[6] https://jesusleal.io/2021/04/21/Longformer-multilabel-classification/

Table 2 The results of accuracy metric in mortality prediction (binary classification) using unsupervised key phrase extraction methods and fine-tuned clinical transformer-based models.

| Row Labels | BERT | PubMedBERT | SciBERT |
|---|---|---|---|
| **Deep Learning** | | | |
| Key2Vec | 0.902 | 0.910 | 0.911 |
| KeyBERT | 0.903 | 0.903 | 0.913 |
| **Graph-Based** | | | |
| MultipartiteRank | 0.905 | 0.912 | 0.915 |
| PositionRank | 0.906 | 0.913 | 0.917 |
| **Statistical** | | | |
| TF-IDF-02 | **0.908** | **0.928** | **0.937** |
| YAKE | 0.905 | 0.923 | 0.925 |
| **Without KE** | | | |
| FullConcepts | 0.903 | 0.916 | 0.918 |

In this experiment about 10% of record has detected as death patients. TF-IDF method for keyphrase extraction in combination with the SciBert representation model lead to best result in binary classification. Table 3 shows the results of F1 and ROC-AUC in diagnosis prediction (multi-label classification). The results of metric F1 in multi-label classification using unsupervised key phrase extraction methods and fine-tuned clinical transformer-based models is shown in Fig 1.

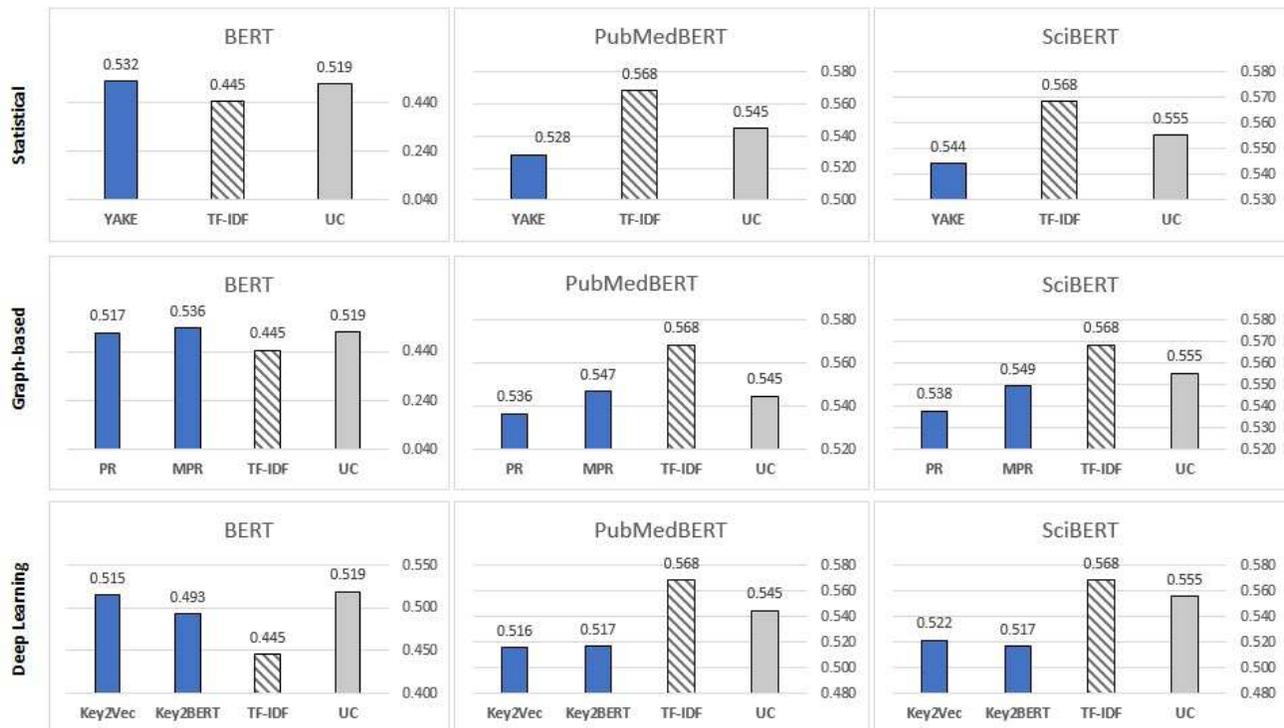

Figure 1 The results of metric F1 in multi-label classification using unsupervised key phrase extraction methods and fine-tuned clinical transformer-based models. The graphs of the first row correspond to statistical method, the graphs of the second row correspond to graph-base method (PR: PositionRank, MPR: MultipartiteRank), and the graphs of the third row correspond to deep learning method. In each column, the effect of using one of the transformer-based models is shown. In each diagram, the bar related to the use of all unique concepts is shown as "UC" and also the bar related to the proposed method is shown as "TF-IDF". The best results are obtained through the proposed method in combination with clinical transformer-based models (PubMedBERT and SciBERT). And the performance of two models PubMedBERT and SciBERT in the metric F1 is almost similar.

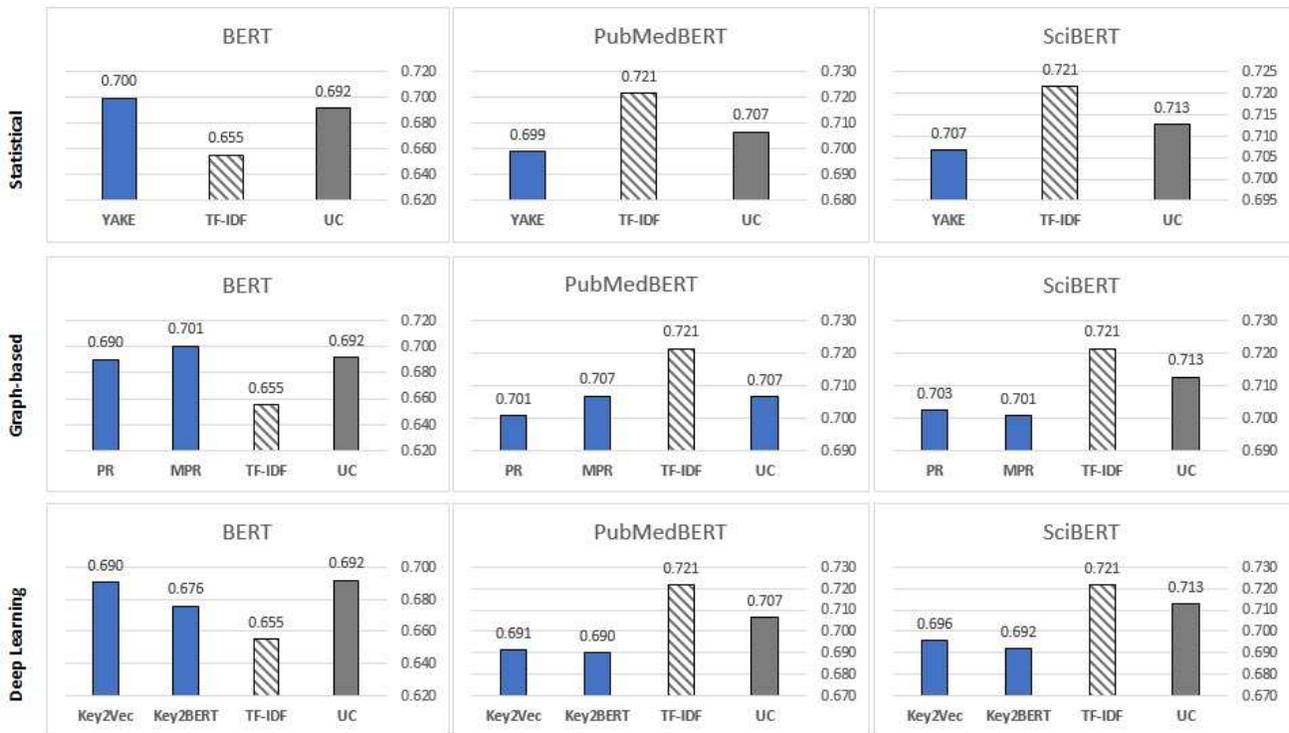

Figure 2 The results of metric ROC-AUC in multi-label classification using unsupervised key phrase extraction methods and fine-tuned clinical transformer-based models. The graphs of the first row correspond to statistical method, the graphs of the second row correspond to graph-base method (PR: PositionRank, MPR: MultipartiteRank), and the graphs of the third row correspond to deep learning method. In each column, the effect of using one of the transformer-based models is shown. In each diagram, the bar related to the use of all unique concepts is shown as "UC" and also the bar related to the proposed method is shown as "TF-IDF". For metric ROC-AUC, we see the superiority of combining the proposed method with clinical transformer-based models (PubMedBERT and SciBERT).

The performance comparison of all input datasets in combination with transformer-based models is shown in Figure 3. The results show the best F1 results were obtained by combining statistical keyphrase extraction with the SciBERT representation model. Furthermore, regardless of the type of representation model, deep learning keyphrase extraction methods values are lower than those of all unique concepts "UC".

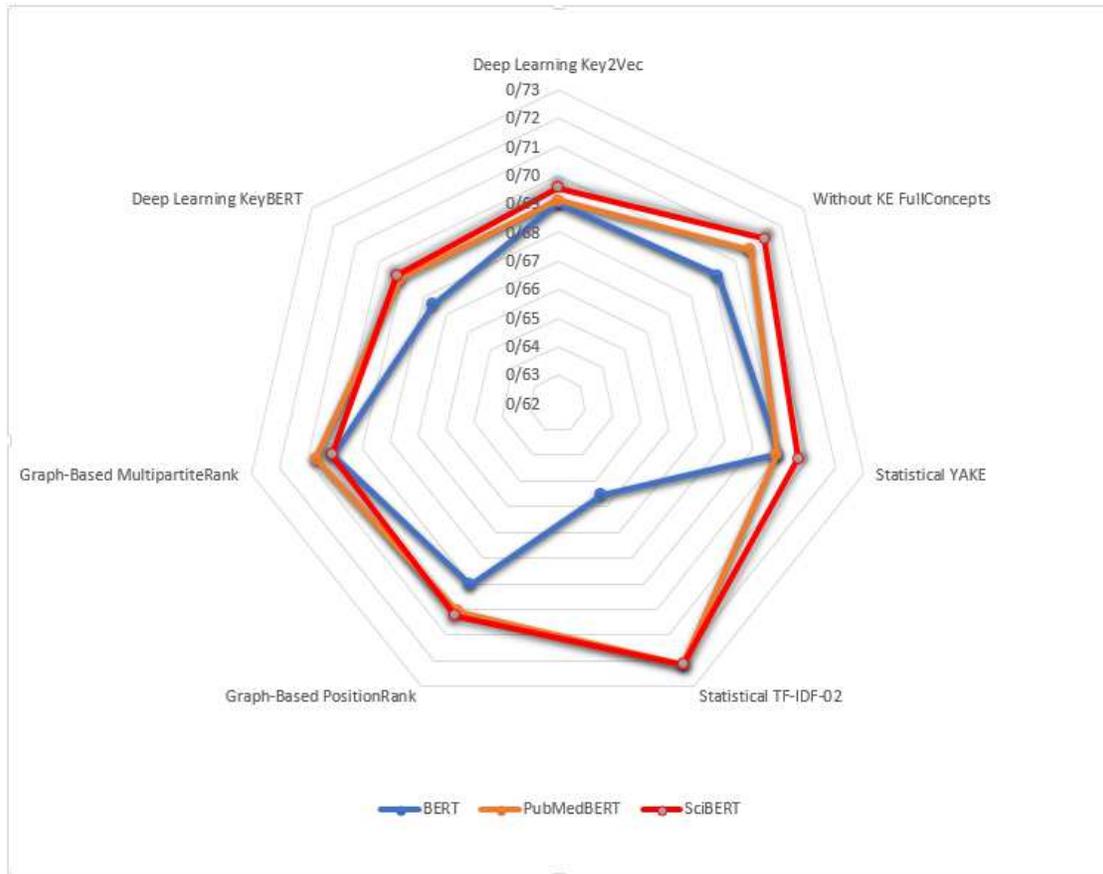

Figure 3- The performance comparison of all input datasets in combination with transformer-based models.

## 5 Discussion

Using machine learning to classify medical text with multi-labels can benefit patient care. Since the clinical notes are an essential part of the medical record, the effort to accurately analyze them can effectively develop primary and secondary EMR data usage. In recent years, the development of language models that are pre-trained on specific clinical data has improved clinical text processing methods.

In [44] and [45], the superiority of transformer-based text representation models over traditional models have been pointed out. Bio-BERT, PubMed-BERT, and UMLS-BERT models have produced better results when receiving a subset of standard concepts extracted from the text. This finding is proof of the vital application of NER and Entity linking.

The variety of writing methods and different levels of expertise of the people completing the clinical texts make mapping clinical terms to standard concepts inevitable. However, a fundamental challenge in using NER and Entity linking is the production of unrelated and low-value concepts [14]. This article focuses on the combination of keyphrase extraction methods and transformer-based text representation methods.

In the presented experiments, it is observed that on all the samples in the dataset, statistical keyphrase extraction methods and the TF-IDF method obtained better results in F1 and ROC-AUC criteria.

### 5.1 Comparison with other studies

In [46], different keyword extraction methods were employed on datasets from multiple languages. This method of evaluation is referred to as partial match evaluation. In this study, the KeyBERT method produces better results, especially on long-term documents, because it preserves the contextual similarity between terms more than other methods. In the current study, the KeyBERT method was not superior at preserving the preferred names of UMLS concepts.

Sun et al. [47] divide unsupervised keyword extraction methods into statistical, graph-based, and topic-based methods. TF-IDF and PositionRank methods outperformed each other on different databases in this study. Both methods use statistical information to identify the key concepts in the text, but the PositionRank method uses the graphic structure to display the document's structure. In our recent study, we have seen these results repeated in a different order, which means that statistical features alone have led to better results. Gero et al. [48], developed a method called NamedKeys. By performing NER, converting multi-word phrases to single words, and then using non-contextual representation methods including word2vec, key concepts are extracted. As contextual representation methods, Transformer-based methods require the addition of new phrases to the vocabulary so that multi-word phrases are converted into single-word phrases. The superiority of contextual embedding methods trained on specific domains for extracting keywords mentioned in [49]. Among the investigated models, the SciBERT and BioBert models have been superior to other methods due to training in scientific articles. This finding confirms the results obtained in this current research.

In this regard, it seems that the main reason for the decrease in efficiency of machine learning-based methods, especially KeyBERT, compared to statistical methods, when receiving input consisting of multi-word phrases corresponding to specialized clinical concepts, is the breaking of the phrase structure during the process of selecting key phrases, which happens because the proper n-gram not specified and grammatical concept structures are not considered at all. In contrast there is no tokenization process in the proposed TF-IDF model. The calculation of the TF-IDF score of the phrases is done based on the code of the phrases, and at the end, the full text of the phrase is mapped to the selected code. But in other methods that have used the text of the concepts as input, we see the generated expression are not always meaningful and there are slightly off-key. Tokenization of multi-word and generating different n-gram without considering concept structure can lead to losing the phrase's meaning. It is worth noting that changing the value of the n-gram cannot eliminate this adverse effect. The outputs produced by different methods for the information of patient number 123103 are listed in Table 3. Most of the generated phrases is meaningless. Pay attention to how Nicotinamide Adenine Dinucleotide (NAD) with (CUI C0027270), a type of enzyme that plays a role in generating energy, is displayed in different implementations.

Table ۳ The KeyBERT method with different n-grams. In the KeyBERT method, the BERT model is used to select keywords. This model includes vocabulary that is trained on many public documents.

| Method | Output |
|---|---|
| CUIs | C0002893,C0007012,**C0027270**,C0032326,C0013604,C0037197,C0011777,C1264639,C0017887,C1761613,C0700124,C0201950, C0018824,C1272695,C0010068,C0336779,C0023031,C0026266,C0032227,C3275121,C0264956,C0032285,C0439688,C0301362, C0005367,C0010054,C0031039,C0024129 |
| Preferred Names of CUIs | Refractory anemias, carbon dioxide, nicotinamide adenine inucleotide (**nad**), pneumothorax, edema, structure of sinus of valsalva, dexamethasone, date/time, nitroglycerin, conjunctival hyperemia, dilated, cholesterol measurement test, heart valve disease, done (qualifier value), coronary heart disease, machine, lanthanum, mitral valve insufficiency, pleural effusion disorder, one vessel coronary disease, atheroma, pneumonia, atelectatic, bromdimethoxyamphetamine, bicarbonates, coronary arteriosclerosis, pericardial effusion, lung volume measurements, |
| Keybert n-gram=4 | atelectatic bromdimethoxyamphetamine bicarbonates coronary, bromdimethoxyamphetamine bicarbonates coronary, bromdimethoxyamphetamine bicarbonates coronary arteriosclerosis, bicarbonates coronary arteriosclerosis pericardial, bicarbonates coronary arteriosclerosis, coronary heart disease, cholesterol measurement test heart, bicarbonates coronary, coronary disease, qualifier value coronary heart, pneumonia atelectatic bromdimethoxyamphetamine bicarbonates, measurement test heart, heart disease machine lanthanum, cholesterol measurement test, arteriosclerosis pericardial effusion lung, **nad** pneumothorax, valve disease done qualifier, **nad pneumothorax edema**, cholesterol measurement, heart disease, |
| Keybert n-gram=3 | bromdimethoxyamphetamine bicarbonates coronary, bicarbonates coronary arteriosclerosis, coronary heart disease, bicarbonates coronary, coronary disease, measurement test heart, cholesterol measurement test, **nad** pneumothorax, **nad** pneumothorax edema, cholesterol measurement, heart disease, coronary arteriosclerosis pericardial, pericardial effusion lung, qualifier value coronary, coronary heart, coronary arteriosclerosis, arteriosclerosis pericardial, disease done qualifier, arteriosclerosis pericardial effusion, pericardial effusion, |
| Keybert n-gram=2 | bicarbonates coronary, coronary disease, **nad** pneumothorax, cholesterol measurement, heart disease, coronary heart, coronary arteriosclerosis, arteriosclerosis pericardial, pericardial effusion, test heart, bromdimethoxyamphetamine bicarbonates, pericardial, coronary, insufficiency pleural, cholesterol, pneumonia atelectatic, heart valve, effusion lung, pneumothorax, arteriosclerosis, |
| Keybert n-gram=1 | pericardial, coronary, cholesterol, pneumothorax, arteriosclerosis, nicotinamide, bicarbonates, anemias, bromdimethoxyamphetamine, nitroglycerin, lung, mitral, pleural, hyperemia, lanthanum, edema, insufficiency, pneumonia, atelectatic, disease, |

For future work we will work on KeyphraseVectorizers in addition to KeyBERT to extract enhanced keyphrases from text documents [50].

In order to verify why the SciBERT model has a better performance than other transformer-based models, we tokenized several specialized words such as the names of diseases by the BERT model and some of other transformer-based language models pre-processed based on clinical data. The obtained result for one expression ("coronary arteriosclerosis") is shown in the Table 4.

Table 4- Comparison of Token for "coronary arteriosclerosis" in vocabularies used by the standard BERT, SciBERT and PubMedBERT.

| Tokenizer | Vocabulary | Pre-trian | Token for "coronary arteriosclerosis" |
|---|---|---|---|
| BERT | BERT-BASE | | ['corona', '##ry', 'arte', '##rio', '##sc', '##ler', '##osis'] |
| SciBERT | SciVocab | Papers from the biomedical domain and computer science | ['coronary', 'arterios', '##cle', '##rosis'] |
| Bio+Clinical | BioBERT BERT-BASE | All MIMIC III, Only the discharge summaries in MIMIC III | ['co', '##rona', '##ry', 'art', '##eri', '##os', '##cle', '##rosis'] |
| BLUEBERT | BERT-BASE | PubMed abstracts, clinical notes from the MIMIC III dataset | ['corona', '##ry', 'arte', '##rio', '##sc', '##ler', '##osis'] |
| PubMed-BERT | BERT-BASE | PubMed (abstracts and full biomedical articles) (3.1B words) | ['coronary', 'arteri', '##osclerosis'] |
| UMLS-BERT | Bio+Clinical-BERT | Patient notes and diagnostic test reports from the MIMIC III | ['co', '##rona', '##ry', 'art', '##eri', '##os', '##cle', '##rosis'] |

The tokenization method in different models is not similar to each other and this issue goes back to their vocabularies. The biomedical term "Coronary" appears in the corresponding vocabulary of The SciBERT and PubMed-BERT but this term will be broken into word pieces in other model forexample in BERT and BLUEBERT, this term broken to "Corona" that is a different and famous disease. These word pieces often have no biomedical relevance and may hinder learning in downstream tasks.

In other words, it can be concluded that despite the progress achieved in contextual models, in order to benefit from NER and linking, the aforementioned models need to be trained on domains that recognize specialized terms. The implementation of this idea will be one of the future works.

This study has some limitations. The first limitation is related to UMLS. The UMLS assignment with standard tools without filtering is error prone due to overlapping. It is also necessary to investigate more representative models especially some pertained transformers that have seen the MIMIC-III texts, e.g. BlueBERT[51] and UmlsBERT[31].

Another limitation is related to percent of death patients in dataset. The mortality of patients in the MIMIC-III database is 23.2%. Older patients are more likely to be admitted to the ICU and have a higher mortality rate. According to the current dataset, a subset of MIMIC-III, about 10% of patients die. In general health centers, this rate is lower. For this reason, it is important to consider this feature when generalizing to another dataset.

# 6 Conclusion

A massive volume of clinical data is generated daily due to the widespread use of EMRs. Using EMR-generated data has become more cost-effective given the improvements in artificial intelligence and processing resources. Since the data recorded in clinical notes could be influenced by factors such as physicians' knowledge and competence, extracting key concepts from clinical notes and mapping them with standard terminologies can be a significant step toward identifying similar cases. However, many concepts extracted from clinical notes are irrelevant and useless for text analysis. The findings of this study show that the effective use of the implementation of processes NER and Entity linking will require a post-processing stage in order to find more valuable concepts. In this study we proposed a simple way extracting key UMLS concepts. The integration of this method with clinical transformer-based model lead to better result in comparison with different category of unsupervised keyphrase extraction methods.

**Data availability**

The datasets generated during and/or analyzed during the current study are available in the
https://physionet.org/content/mimiciii/1.4/.
Codes are available at https://github.com/HodaMemar/A3.